\pdfoutput=1

\documentclass[11pt]{article}

\usepackage[final]{EMNLP2022}

\usepackage{times}
\usepackage{latexsym}

\usepackage[T1]{fontenc}

\usepackage[utf8]{inputenc}

\usepackage{microtype}
\usepackage{inconsolata}

\usepackage{xspace}

\usepackage{todonotes}

\newcommand{\ours}{\textsc{ConQRR}\xspace} 

\usepackage{graphicx}
\usepackage{amssymb}
\usepackage{amsmath}
\usepackage{bbm}
\DeclareMathOperator*{\argmax}{arg\,max}

\usepackage{esvect}
\usepackage{svg}
\usepackage{comment}

\usepackage{comment}
\usepackage{subcaption}
\usepackage{graphicx}

\usepackage{booktabs, multirow} 
\usepackage{soul}
\usepackage{makecell}
\usepackage{threeparttable}
\usepackage{footnote}
\makesavenoteenv{tabular}

\title{\ours: Conversational Query Rewriting for Retrieval with Reinforcement Learning
}

\newcommand\uw{$^{\diamondsuit}$}
\newcommand\google{$^\spadesuit$}
\newcommand\aitwo{$^{\clubsuit}$}
\newcommand\first{$^\ast$}
\newcommand\aspace{\hspace{.75em}}

\author{
    Zeqiu Wu \uw\thanks{\first Work done during an internship at Google Research.}\aspace
    Yi Luan \google\aspace
    Hannah Rashkin \google\aspace
    David Reitter \google\aspace \\
    \textbf{Hannaneh Hajishirzi} \uw\aitwo\aspace
    \textbf{Mari Ostendorf} \uw\aspace
    \textbf{Gaurav Singh Tomar} \google\\
    \uw University of Washington \aspace
    \google Google Research \aspace
    \aitwo Allen Institute for AI\\
    {\tt \{zeqiuwu1,hannaneh,ostendor\}@uw.edu} \\
    {\tt \{luanyi,hrashkin,reitter,gtomar\}@google.com}
}

\begin{document}
\maketitle
\begin{abstract}
Compared to standard retrieval tasks, passage retrieval for conversational question answering (CQA) poses new challenges in understanding the current user question, as each question needs to be interpreted within the dialogue context.
Moreover, it can be expensive to re-train well-established retrievers such as search engines that are originally developed for non-conversational queries. To facilitate their use, we develop a query rewriting model \ours that rewrites a conversational question in the context into a standalone question. It is trained with a novel reward function to directly optimize towards retrieval using reinforcement learning and can be adapted to any off-the-shelf retriever. \ours achieves state-of-the-art results on a recent open-domain CQA dataset containing conversations from three different sources, and is effective for two different off-the-shelf retrievers.
Our extensive analysis also shows the robustness of \ours to out-of-domain dialogues as well as to zero query rewriting supervision.
\end{abstract}

\section{Introduction}
\label{sec:intro}

Passage retrieval in an open-domain conversational question answering (CQA) system \citep{anantha-etal-2021-open}, compared to standard retrieval tasks \citep{voorhees-tice-2000-trec, nguyen2016ms}, poses new challenges of understanding user questions within the dialogue context.
Most existing conversational retrieval models \citep{yu2021convdr, lin-etal-2021-contextualized, kim2022saving} rely on training specific retrievers like dual encoders \citep{karpukhin-etal-2020-dense}.
However, re-training well-established retrievers for conversational queries can be expensive or even infeasible due to their complicated system designs (e.g., those used in search engines). Moreover, the preference and availability of such off-the-shelf retrievers can vary depending on the end users. 

The task of question-in-context rewriting or query rewriting (QR) in a conversation \citep{elgohary-etal-2019-unpack, dalton-2019-cast} is to convert a context-dependent question into a self-contained question. It enables the use of any off-the-shelf retriever (Table~\ref{tab:compare_related}), which we define as a retriever that cannot be fine-tuned or provide access to any of its internal architecture design or intermediate results (i.e., can only be seen as a black-box).

\begin{figure}
  \centering
  \includegraphics[width=0.45\textwidth]{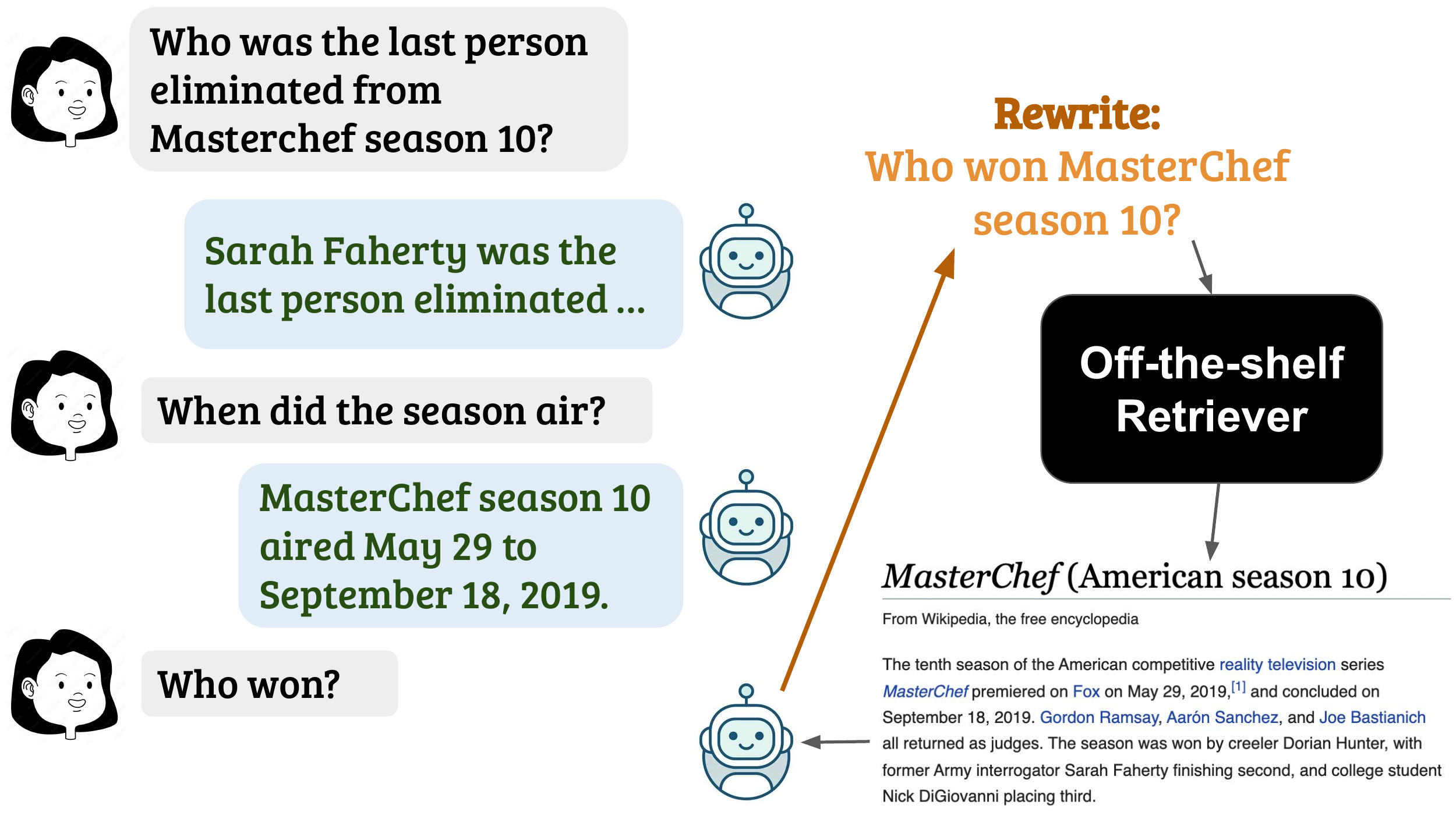}
  \caption{A CQA agent rewrites the current user question into a more effective one (in orange) for the given \textit{off-the-shelf} retriever to find the most relevant passage.}
 \label{fig:teaser}
\end{figure}

Therefore, in this paper, we focus on \textit{query rewriting} for the task of \emph{conversational passage retrieval} in a CQA dialogue with \textit{any off-the-shelf} retrieval system that can only be used as a black box. Specifically, we seek to build a QR model that rewrites a user query into the input of the retriever, in such a way that optimizes for passage retrieval performance.  Figure~\ref{fig:teaser} shows an example of our task, where given an off-the-shelf retriever, the agent rewrites the current user query ``Who won?'' into a more effective query 
for retrieval.

Recent work that leverages QR for conversational passage retrieval \cite{anantha-etal-2021-open, dalton-2019-cast} collects human-rewritten queries to train a supervised QR model. However, humans are usually instructed to rewrite conversational queries to be unambiguous to a human outside the dialogue context, which does not necessarily align with \textit{the goal in our task}---to optimize the retrieval performance. We conduct comprehensive experiments in Section~\ref{sec:analysis} to confirm these human rewrites indeed sometimes omit information from the dialogue context that is useful to the retriever. This limitation of human query rewrites impacts supervised training. In addition, prior supervised QR models are agnostic to downstream retrievers as they are separately trained before their predicted rewrites being used for retrieval during inference.

We propose a reinforcement learning (RL)-based model \ours (\textbf{Con}versational \textbf{Q}uery \textbf{R}ewriting for \textbf{R}etrieval). It directly optimizes the rewritten query towards retrieval performance, using only weak supervision from retrieval. 
We adopt a novel reward function that computes an approximate but effective retrieval performance metric on in-batch passages at each training step. Our reward function does not assume any specific retriever model design, and is generic enough for \ours to adapt to any off-the-shelf retriever. 


We show \ours outperforms existing QR models on a recent large-scale open-domain CQA dataset QReCC \cite{anantha-etal-2021-open} by over 12\% and 14\% for BM25 and a neural dual encoder retriever model \cite{ni-t5x} 
respectively, averaging over three retrieval metrics. We observe the performance boost on all three QReCC subsets from different conversation sources, including one that only appears in the test set (i.e., out-of-domain). 

To conclude, our contributions are as follows. 1) We introduce \ours as the first RL-based QR model that can be adapted to and optimized towards any off-the-shelf retriever for conversational retrieval. 2) We demonstrate that \ours achieves state-of-the-art results with off-the-shelf retrievers on QReCC with conversations from three sources, and is effective for two retrievers including BM25 and a dual encoder model. 3) Our analysis shows \ours trained with \textit{no human rewrite supervision} provides better retrieval results than strong baselines trained with full supervision, and is robust to out-of-domain dialogues, topic shifts and long dialogue contexts. 4) We conduct a novel quantitative study to analyze the limitations and utility of human rewrites in retrieval performance, which are largely unexplored in prior work.


\section{Related Work}
\label{sec:related}

\subsection{Conversational Question Answering}
Most existing CQA datasets \cite{choi-etal-2018-quac, reddy-etal-2019-coqa} are designed for the task of reading a document to answer questions in a conversation, which does not require the retrieval step. In contrast, QReCC \citep{anantha-etal-2021-open} is a recent open-domain CQA dataset where a conversational agent retrieves the most relevant passage(s) before generating an answer to the question.

\begin{table}
\normalsize
\centering
\resizebox{0.48\textwidth}{!}{
\begin{tabular}{l|cc}\toprule
&Fine-Tune &Arch Type\\\midrule
ConvDR \citep{yu2021convdr} & Part & Dual Encoder\\
CQE \citep{lin-etal-2021-contextualized} & Part & Dual Encoder\\
\citet{kim2022saving} & Yes & Dual Encoder\\
\midrule
\textbf{QR for Retrieval} & \textit{No} &\textit{No Limit}\\
\bottomrule
\end{tabular}
}
\caption{Retriever requirements of different frameworks for conversational retrieval. \textit{Arch Type} stands for \textit{Retriever Architecture Type}. 
}
\label{tab:compare_related}
\vspace{-2mm}
\end{table}
\vspace{-1mm}
\subsection{Conversational Retrieval} 
A few recent works \citep{dalton-2019-cast, qu-2020-orquac} collect retrieval datasets for \textit{conversational search} tasks \citep{BELKIN1995379, SOLOMON1997217} where each dialogue context consists of a sequence of previous user questions only. \citet{dalton-2019-cast} annotate 80 conversations for the TREC CAsT-19 task and \citet{qu-2020-orquac} derive their dataset based on QuAC \citep{choi-etal-2018-quac} and propose to fine-tune a dual encoder retriever \citep{guu2020realm, karpukhin-etal-2020-dense}. In contrast, the dialogue context in a CQA conversation, which is the focus of our work, consists of both user and agent turns. Each user query in a CQA conversation can be more challenging to de-contextualize as it depends on both previous user and agent turns. 

Most existing conversational retrieval models require fine-tuning a retriever of a specific type (Table~\ref{tab:compare_related}). \citet{yu2021convdr}, \citet{lin-etal-2021-contextualized} and \citet{kim2022saving} attempt to fine-tune a dual encoder retriever \citep{xiong2021approximate, karpukhin-etal-2020-dense} to handle conversational queries. \citet{kumar-callan-2020-making} propose a framework focusing on improving the passage re-ranker after the retrieval. 

\vspace{-1mm}
\paragraph{Query Rewriting (QR)}

In order to directly use an \textit{off-the-shelf} retriever as we aim to do, conversational QR \citep{elgohary-etal-2019-unpack} has been applied in prior work \citep{Vakulenko2021trans++, lin2020t5qr, yu-2020-few-shot-ir,Voskarides_2020} to first convert a conversational query into a standalone one. 
\citet{yu-2020-few-shot-ir} propose a supervised QR model trained with human rewrites and weak QR supervisions specifically for conversational search tasks that are generated from additional search session resources.
\citet{lin2020t5qr} and \citet{Vakulenko2021trans++} also use human rewrites to train a supervised QR model based on pre-trained language models like T5 \citep{raffel2020exploring} or GPT2 \citep{Radford2019gpt2}. \citet{Voskarides_2020} use human rewrites to train a model that classifies whether each token in the dialogue context should be used to construct the query for retrieval. In contrast, we show the limitations of human rewrites used as QR supervision and design an RL-based QR model which can achieve better performance than supervised models even without human rewrites. Similar to our finding with details in Appendix~\ref{sec:appendix_analysis}, \citet{ishii-etal-2022-question} claim that using rewritten queries as an intermediate step does not necessarily outperform fine-tuning the end task model (e.g., retriever). However, we provide strong evidence in Section~\ref{sec:analysis} to support the importance of QR in the \textit{off-the-shelf} retriever setting.

\subsection{RL for Text Generation} 

\paragraph{RL-based QR for Retrieval} 
\citet{nogueira-cho-2017-task} and \citet{adolphs2021boosting} apply RL based on gold passage labels to do \textit{non-conversational} query reformulation for retrieval. In contrast, to the best of our knowledge, we are the first to apply RL for rewriting \textit{conversational} queries, and we only use weak retrieval supervision and an approximate retrieval metric for computational efficiency. Additionally, our model rewrites the query based on the dialogue context, while their models require multiple rounds of retrieval in order to reformulate a query, which can be time-consuming. 
\vspace{-1mm}
\paragraph{Other Applications}
Prior work also applies RL approaches to address text generation tasks like machine translation \citep{RanzatoCAZ15, wu2016googles}, text summarization \citep{Paulus2018RLSum, Celikyilmaz2018DeepComAgent} and image captioning \cite{rennie2017self-critical, fisch-etal-2020-capwap} by training a model directly optimized towards generation quality metrics like BLEU, ROUGE or CIDEr. \citet{buck2018ask} use RL to rewrite a \textit{non-conversational} query for the task of question answering model.

\section{Approach}
\label{sec:approach}

\paragraph{Problem Definition} We focus on the task of \textit{query rewriting (QR)} for \textit{conversational passage retrieval} in a CQA dialogue, with an \textit{off-the-shelf} retriever. The task inputs include a dialogue context $x$ consisting of a sequence of previous utterances $(u_1, u_2, \dots, u_{n-1})$, the current user question $u_n$, a passage corpus $P$ and an off-the-shelf retriever $R$.\footnote{To mimic practical use cases, $R$ is usually assumed to be general purpose retriever with standard search queries.} $R$ cannot be fine-tuned but returns a ranked list of top-k passages when given a query string and a passage corpus, and no other assumption about the model architecture of $R$ can be made. The task aims to rewrite $x$ into a query $q$ such that $R$ can take $q$ as the input query to retrieve passages relevant to $x$ from $P$. Specifically, a passage $p$ is relevant to $x$ if $p$ provides enough information to answer $u_n$ in the context of $(u_1, u_2, \dots, u_{n-1})$.

In this section, we first describe a supervised QR model based on T5 (T5QR) \citep{lin2020t5qr} that applies a generic Seq2Seq training objective with QR labels (Section~\ref{sec:t5qr}). Then we introduce our RL-based framework \ours (\textbf{Con}versational \textbf{Q}uery \textbf{R}ewriting for \textbf{R}etrieval) that trains a QR model to optimize towards retrieval and is adaptable to any given off-the-shelf retriever, with weak retrieval supervision (Section~\ref{sec:model}).

\subsection{T5QR}
\label{sec:t5qr}
T5 is an encoder-decoder model that is pre-trained on large textual corpora \citep{raffel2020exploring}. Following \citet{lin2020t5qr}, we fine-tune T5 to rewrite a conversational query with the input as the concatenation of utterances in the dialogue context $x$ and the output as the human rewrite $\hat{q}$. Note that we concatenate the utterances in a reversed order such that $u_n$ becomes the first one in the input string and any truncation impacts more distant context. Utterances are separated with a separator token ``[SEP]'' in the concatenated string. The model is then trained with a standard cross entropy (CE) loss to maximize the likelihood of generating $\hat{q}$, which is a self-contained version of the query $u_n$ that can be interpreted without knowing previous turns $(u_1, u_2, \dots, u_{n-1})$ in $x$.

\subsection{\ours}
\label{sec:model}
\begin{figure}
  \centering
  \includegraphics[width=0.45\textwidth]{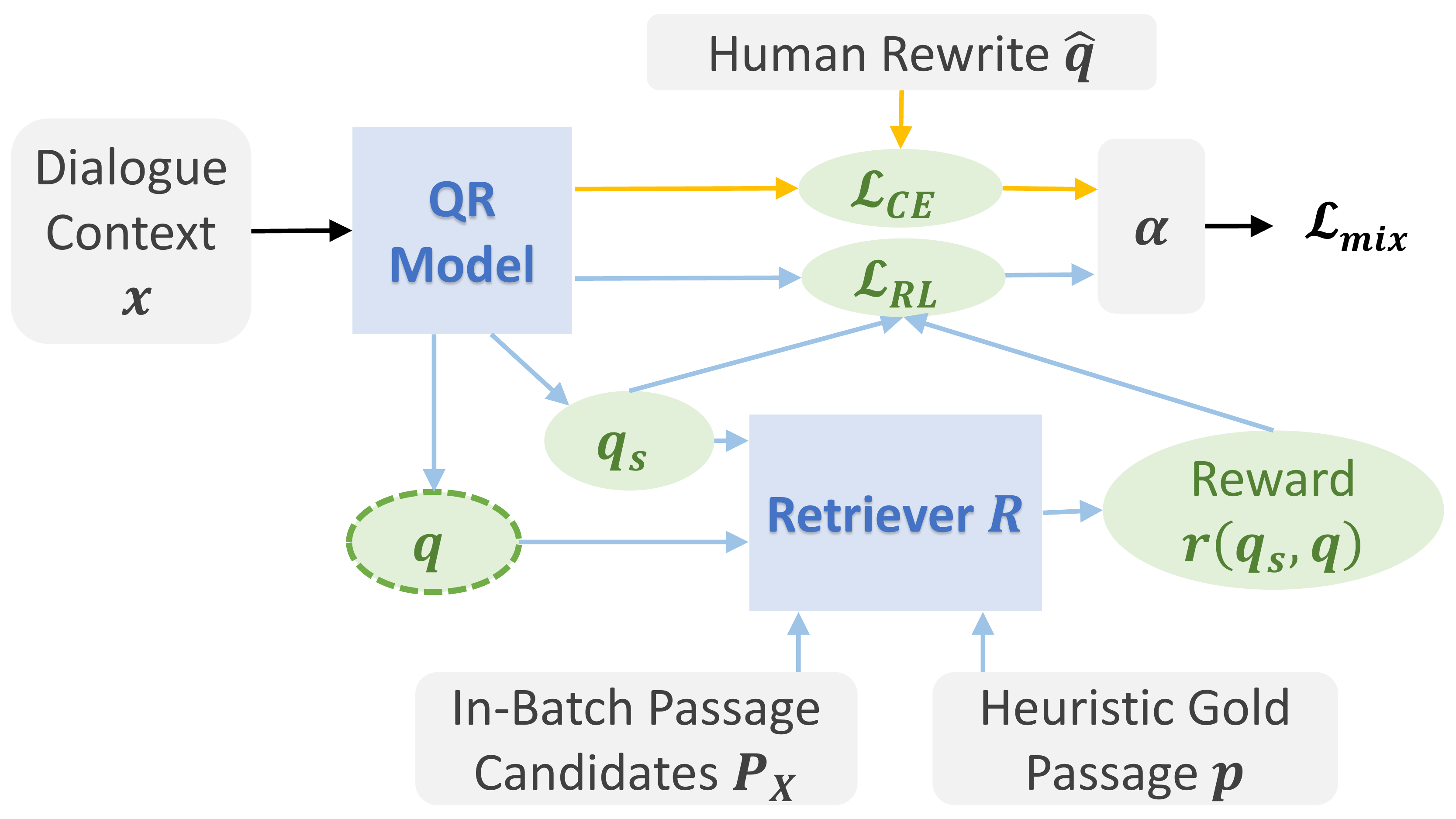}
  \caption{Our \ours framework. Yellow and blue arrows mark the flow of CE (\textbf{unused when $\alpha=1.0$}) and RL loss calculation, respectively. During inference, only $q$ (dashed border) is generated as the final rewrite.
  }
 \label{fig:model}
 \vspace{-2mm}
\end{figure}
QR models trained with a standard CE loss are agnostic to the retriever. In addition, human rewrites are not necessarily the most effective ones for passage retrieval  (see Section~\ref{sec:analysis} for an exploration).

This motivates us to design our RL-based framework \ours (Figure~\ref{fig:model}) that trains a QR model directly optimized for the retrieval performance and can be adapted to any given off-the-shelf retriever. 
Here, the RL environment includes the retriever model, dialogue context and passage candidates, in which the QR model takes actions by generating rewritten queries and obtains rewards accordingly.

To be comparable with supervised QR models that do not use gold passages in training, we first describe how we obtain weak retrieval supervision for the RL reward calculation in \ours. Then we introduce the RL training details of \ours.
\vspace{-1mm}
\paragraph{Weak Retrieval Supervision} 
In a CQA dialogue, each question naturally comes with an answer in its following conversational utterance.
For each $x$, we mark its weak passage label $p$ as the one having a string span with the highest token overlap F1-score with the following answer string $u_{n+1}$:
\begin{equation}
    p = \argmax_{p' \in P} \big[ \argmax_{s \in p'} sim(s, u_{n+1}) \big]
\label{eq:heuristic}
\end{equation}
where $s$ is a string span and $sim()$ calculates the token overlap score between two strings.\footnote{We randomly choose a passage if there is a tie in scores.} Tokens are lower-cased from the NLTK tokenizer.\footnote{\url{https://www.nltk.org}} However, as searching within all candidates in $P$ is very time-consuming, we instead first use BM25 to retrieve the top 100 passages from $P$ with the BM25 input being the human rewrite,\footnote{We show in Section~\ref{sec:analysis} that using the dialogue context as the BM25 input to induce weak supervision gives similar performance (Figure~\ref{fig:data_efficiency}), where no human rewrites are used.} and then locate the best passage $p$ from these 100 candidates.

\vspace{-1mm}
\paragraph{RL Training}
\ours also has T5 as the base model architecture. 
It can be initialized with either T5 or T5QR. Our analysis in Section~\ref{sec:exp} shows that both setups generally work well.

For each training example with the dialogue context $x$, we use the concatenated utterances in $x$ as the model input. For each input, we generate $m$ sampled rewritten queries $(q_{s_1}, \dots, q_{s_m})$ as well as a baseline generated rewrite $q$. To generate each sampled rewrite $q_s$, at time step $t$ of the decoding process, a token $q_{s}^t$ is drawn from the decoder probability distribution $Pr(w|x,q_{s}^{1:t-1})$ 
The baseline rewrite $q$ is the output of greedy decoding,\footnote{We tried beam search with various beam sizes and got similar results as greedy decoding.} which is also applied for query rewriting during inference. We then apply a self-critical sequence training algorithm \citep{rennie2017self-critical} to calculate the reward for each $q_s$ relative to $q$ as $r(q_s, q) = score(q_s) - score(q)$. The intuition is to reward/penalize the generation of sampled rewrites that lead to better/worse retrieval performance than greedy decoding used during inference. Ideally, the $score()$ function should be some retrieval evaluation metric like mean reciprocal rank (MRR) or Recall@K. However, as it is very costly to run actual retrieval for each training step, we instead use an approximate scoring function described below. 

To compute $score(q)$ for a rewrite $q$, we first use $q$ to do retrieval from the in-batch passage candidates $P_X$ defined as follows, instead of from the full passage corpus $P$. We pre-compute one positive and one hard negative passage ($p$ and $p_n$) for each training example $x$ where $p_n$ is a randomly selected passage that is different from $p$, 50\% of the time from the top 100 BM25-retrieved candidates (with the BM25 input being the human rewrite) and remaining 50\% of the time from $P$. We define the set of all such positive and negative passages of input examples in a batch $X$ as the in-batch passage candidates $P_X$. Formally, we define $P_X=\{p^i, p_n^i| x_i \in X\}$ as the set of in-batch passage candidates for the batch $X$. Then for a generated rewritten query $q$ of $x\in X$, we calculate $score(q)$ as a binary indicator of whether the retriever R ranks the assigned positive passage $p$ highest from $P_X$. We denote $R(q, P_X, k)$ as the $k$-th most relevant passage retrieved by $R$ from the candidate pool $P_X$, and define:
\begin{equation}
score(q) = \mathbbm{1} \big[R(q, P_X, 1) = p\big]
\label{eq:score}
\end{equation}
Then the RL training loss for $x$ becomes: 
\begin{align*}
&\mathcal{L}_{RL} = -\frac{1}{m} \sum_{i=1}^{m} r(q_{s_i}, q)\log Pr(q_{s_i}|x) \\
&Pr(q_{s_i}|x) = \prod_{t=1}^{|q_{s_i}|} Pr(q_{s_i}^t|x, q_{s_i}^{1:t-1})
\end{align*}
Following prior work \citep{Paulus2018RLSum, Celikyilmaz2018DeepComAgent}, we experiment with a pure RL loss ($\mathcal{L}_{RL}$) and a mixed RL and CE loss in training: 
\begin{equation}
\mathcal{L}_{mix} = \alpha \mathcal{L}_{RL} + (1-\alpha) \mathcal{L}_{CE}
\label{eq:loss}
\end{equation}
where $\alpha\in [0,1]$ is a tunable parameter.
\vspace{-1mm}
\paragraph{Inference}
At inference time, both T5QR and \ours work in the same way. The trained QR model greedily generates the rewritten query given a dialogue context. Then, the predicted rewrite is given to the provided retriever to perform retrieval.

\subsection{Retriever Models}
\label{sec:retrievers}
We evaluate the effectiveness of \ours in experiments with two general-domain retrieval systems, with more details in Appendix~\ref{sec:appendix_impl}.
\vspace{-1mm}
\paragraph{BM25} We follow \citet{anantha-etal-2021-open} using Pyserini \citep{yang-anserini} with default parameters $k1=0.82$ and $b=0.68$. 
\vspace{-1mm}
\paragraph{Dual Encoder (DE)} We use a recent T5-base dual encoder model \cite{ni-t5x} which achieves state-of-the-art performance on multiple retrieval benchmarks. This model is fine-tuned on MS MARCO, and kept fixed for our experiments. 
 
\section{Experiment}
\label{sec:exp}


\paragraph{Dataset} QReCC \citep{anantha-etal-2021-open} is a dataset of 14k open-domain English conversations in the format of alternating user questions and agent-provided answers with 80k question and answer pairs in total. The conversations are collected from different sources: QuAC \citep{choi-etal-2018-quac}, Natural Questions \citep{kwiatkowski-etal-2019-natural} and TREC CAsT-19 \citep{dalton-2019-cast} with additional annotations by crowd workers. See more details and statistics in Appendix~\ref{sec:appendix_data}. Therefore, QReCC can be divided into three subsets for evaluation. We name them as \textit{QuAC-Conv}, \textit{NQ-Conv} and \textit{TREC-Conv} respectively to differentiate them from the original datasets from which they are derived. 
TREC-Conv only appears in the test set. Each user question comes with a human-rewritten query. For each agent turn, gold passage labels are provided if any. The entire text corpus for retrieval contains 54M passages, segmented in the released data.\footnote{Original QReCC data: \url{https://zenodo.org/record/5115890\#.YZ8kab3MI-Q}.}

\paragraph{Evaluation Metrics} Following \cite{anantha-etal-2021-open}, we use mean reciprocal rank (MRR), Recall@10 and Recall@100 to evaluate the retrieval performance by using the provided evaluation scripts.\footnote{Both original and updated evaluation scripts: \url{https://github.com/scai-conf/SCAI-QReCC-21}.}
We use their \textit{updated} evaluation script for most experiments, except that we also use the \textit{original} version for calculating scores in Table~\ref{tab:comparable_eval} to compare with their reported QReCC baseline results. We note that these two evaluation scripts only differ by a scaling factor\footnote{This is due to the exclusion of test examples with no valid gold passage labels (roughly 50\%) in the updated evaluation, which results in 6396, 1442 and 371 test instances for QuAC-Conv, NQ-Conv and TREC-Conv, respectively.} so they should lead to the same conclusions regarding model comparisons. See more details in Appendix~\ref{sec:appendix_eval}.

\begin{table}
\normalsize
\centering
\resizebox{0.48\textwidth}{!}{
\begin{tabular}{l|ccc|ccc}\toprule
\multirow{3}{*}{\textbf{QR Model}}
&\multicolumn{3}{c|}{\textbf{Original Eval}} &\multicolumn{3}{c}{\textbf{Updated Eval}}\\
&MRR &R10 &R100 &MRR &R10 &R100 \\\midrule
GPT2 + WS & 0.152 & 24.7 & 41.5 & 0.304 & 49.6 & 83.1 \\
Transformer++ & 0.155 & 24.8 & 40.6 & 0.311 & 49.8 & 81.4 \\
T5QR & 0.164 & 26.2 & 42.3 & 0.328 & 52.5 & 84.7 \\
\ours (mix) & 0.186 & 29.2 & \textbf{45.0} & 0.373 & 58.5 & \textbf{90.2} \\
\ours (RL) & \textbf{0.191} & \textbf{30.0} & 44.4 & \textbf{0.383} & \textbf{60.1} & 88.9 \\\midrule
Human & 0.199 & 32.8 & 49.4 & 0.398 & 62.6 & 98.5 \\
\bottomrule
\end{tabular}
}
\caption{Passage retrieval performance of QR models, comparable to scores in \citet{anantha-etal-2021-open} by using the same BM25 retriever for QReCC test set. \ours achieves \textit{state-of-the-art} results. Recall@10 and Recall@100 are abbreviated as R10 and R100. 
}
\label{tab:comparable_eval}
\end{table}
\begin{table*}
\centering
\resizebox{0.97\textwidth}{!}{
\begin{tabular}{ll|ccc|ccc|ccc|cccc}\toprule
\multirow{3}{*}{\textbf{QR Model}}
&\multirow{3}{*}{\textbf{IR System}}
&\multicolumn{3}{c|}{\textbf{QReCC (Overall)}} &\multicolumn{3}{c}{\textbf{QuAC-Conv}}
&\multicolumn{3}{c}{\textbf{NQ-Conv}} &\multicolumn{3}{c}{\textbf{TREC-Conv (OOD)}$^*$}\\
&&MRR &R10 &R100 &MRR &R10 &R100 &MRR &R10 &R100 &MRR &R10 &R100 \\\midrule
T5QR & BM25  & 0.328 & 52.5 & 84.7 & 0.33 & 52.7 & 85.0 & 0.345 & 54.2 & 83.9 & \textbf{0.230} & 44.5 & 82.3 \\
\ours (mix) & BM25 & 0.373 & 58.5 & \textbf{90.2} & 0.379 & 59.2 & \textbf{90.9} & \textbf{0.385} & \textbf{58.8} & \textbf{88.9} & 0.229 & \textbf{44.7} & \textbf{82.7} \\
\ours (RL) & BM25  & \textbf{0.383} & \textbf{60.1} & 88.9 & \textbf{0.395} & \textbf{61.6} & 90.2 & 0.378 & 58.0 & 86.7 & 0.198 & 43.5 & 75.9 \\\midrule
Human Rewrite & BM25    & 0.398 & 62.6 & 98.5 & 0.403 & 62.9 & 98.4 & 0.408 & 63.8 & 99.0 & 0.273 & 53.8 & 98.9 \\\midrule\midrule
T5QR & DE & 0.361 & 56.2 & 75.9 & 0.349 & 55.7 & 76.1 & 0.417 & 58.7 & 74.2 & 0.343 & 55.9 & 79.2 \\
\ours (mix) & DE & 0.395 & 61.9	& 81.8 & 0.387 & 62.0 & 82.4 & 0.439 & 62.2 & 79.0 & \textbf{0.361} & \textbf{58.9} & \textbf{81.0} \\
\ours (RL) & DE & \textbf{0.418} & \textbf{65.1} & \textbf{84.7} & \textbf{0.416} & \textbf{65.9} & \textbf{85.8} & \textbf{0.453} & \textbf{64.1} & \textbf{80.9} & 0.327 & 55.2 & 79.6 \\\midrule
Human Rewrite & DE      & 0.422 & 64.8 & 84.0 & 0.409 & 64.5 & 84.1 & 0.483 & 65.8 & 83.2 & 0.411 & 66.0 & 86.5 \\
\bottomrule
\end{tabular}
}
\caption{Passage retrieval performance on QReCC test set and 3 subsets. \ours (mix) beats the supervised T5QR model on all retriever system and test set combinations. $^*$ OOD (out-of-domain): only appear in the test set. 
}
\label{tab:main_results}
\end{table*}

\paragraph{Implementation Details}
Following prior work on RL for text generation \citep{Paulus2018RLSum, fisch-etal-2020-capwap}, we first initialize \ours with a supervised model (T5QR) \citep{lin2020t5qr} as a warm-up. Our RL optimization (self-critical sequence training \citep{rennie2017self-critical}) uses a policy gradient method with Monte Carlo sampling. In Section~\ref{sec:analysis}, we show that although initializing with T5QR works better than T5, both setups generally work well. All our models use T5-base as the base model. 
We experiment with \ours trained with either a mixed ($\mathcal{L}_{mix}$) or pure RL ($\mathcal{L}_{RL}$) loss. For the mixed loss, we observe that \ours works well when the RL loss weight $\alpha$ is large.\footnote{We also experiment with $\alpha=0.0$, where the RL loss is removed for both retrievers, and get similar results as T5QR.} We tune its values in {0.9, 0.95, 0.97, 0.99}, and use 0.99 as the final value. 
Due to space limit, more implementation and hyper-parameter details are reported in Appendix~\ref{sec:appendix_impl}.

\subsection{Compared Systems}
\label{sec:systems}
For QR models, we compare three supervised models including \textbf{GPT2 with weak supervision (WS)} \citep{yu-2020-few-shot-ir}, a GPT2-medium based system that additionally leverages search sessions to create weak supervision for QR training before fine-tuning, \textbf{T5QR} \citep{lin2020t5qr} and \textbf{Transformer++}, the previous state-of-the-art model based on GPT2-medium \citep{Vakulenko2021trans++} and reported in the original dataset paper \citep{anantha-etal-2021-open}, as well as \textbf{\ours (mix/RL)} with a mixed ($\mathcal{L}_{mix}$) or pure RL ($\mathcal{L}_{RL}$) loss. For analysis purposes, we also report performance for directly using the concatenated dialogue context as the retriever input without any query rewriting in Section~\ref{sec:analysis}. We experiment with two off-the-shelf retrievers, \textbf{BM25} and \textbf{DE} (Section~\ref{sec:retrievers}). 

\subsection{Quantitative Results}
\label{sec:quantitative_results}
To have a direct comparison with the original QR baseline Transformer++, which has the retrieval performance reported on the overall QReCC test set by using BM25 as the off-the-shelf retriever, we first compare all QR models in the same setting in Table~\ref{tab:comparable_eval} and use both the original and updated versions of the provided evaluation script. GPT2 + WS has similar performance as Transformer++. T5QR and \ours outperform the Transformer++ baseline by 5\% and 18\% respectively, averaged on three metrics,\footnote{We obtained prediction results from the authors and reran their evaluation script. The numbers we got are slightly lower than what they reported, but do not affect the conclusions.} although Transformer++ is based on a larger base model - GPT2-medium. Therefore, \ours (RL) becomes the \textit{state-of-the-art} QR model for conversational passage retrieval on QReCC with the original BM25 retriever in \citet{anantha-etal-2021-open}.

Table~\ref{tab:main_results} shows more comprehensive retrieval results comparing \ours and the supervised model T5QR, with the updated evaluation script. 
For the overall QReCC test set, \ours outperforms T5QR for all three metrics. For MRR and Recall@10, gains are roughly 15\% with the RL loss and 9-14\% with the mixed loss for both retrievers. Gains in Recall@100 vary more (4-12\%). Breaking down the results by subset shows that the mixed loss is more robust. \ours (RL) is less effective for the TREC-Conv subset, which only appears in the test set. This suggests that RL loss alone does not generalize well to out-of-domain examples. Across all subsets, the best MRR and Recall@10 results are consistently from DE, whereas BM25 has better Recall@100 scores. See our explanation in Appendix~\ref{sec:appendix_analysis}.

\subsection{Analysis}
\label{sec:analysis}

\begin{figure}
  \centering
  \includegraphics[width=0.4\textwidth]{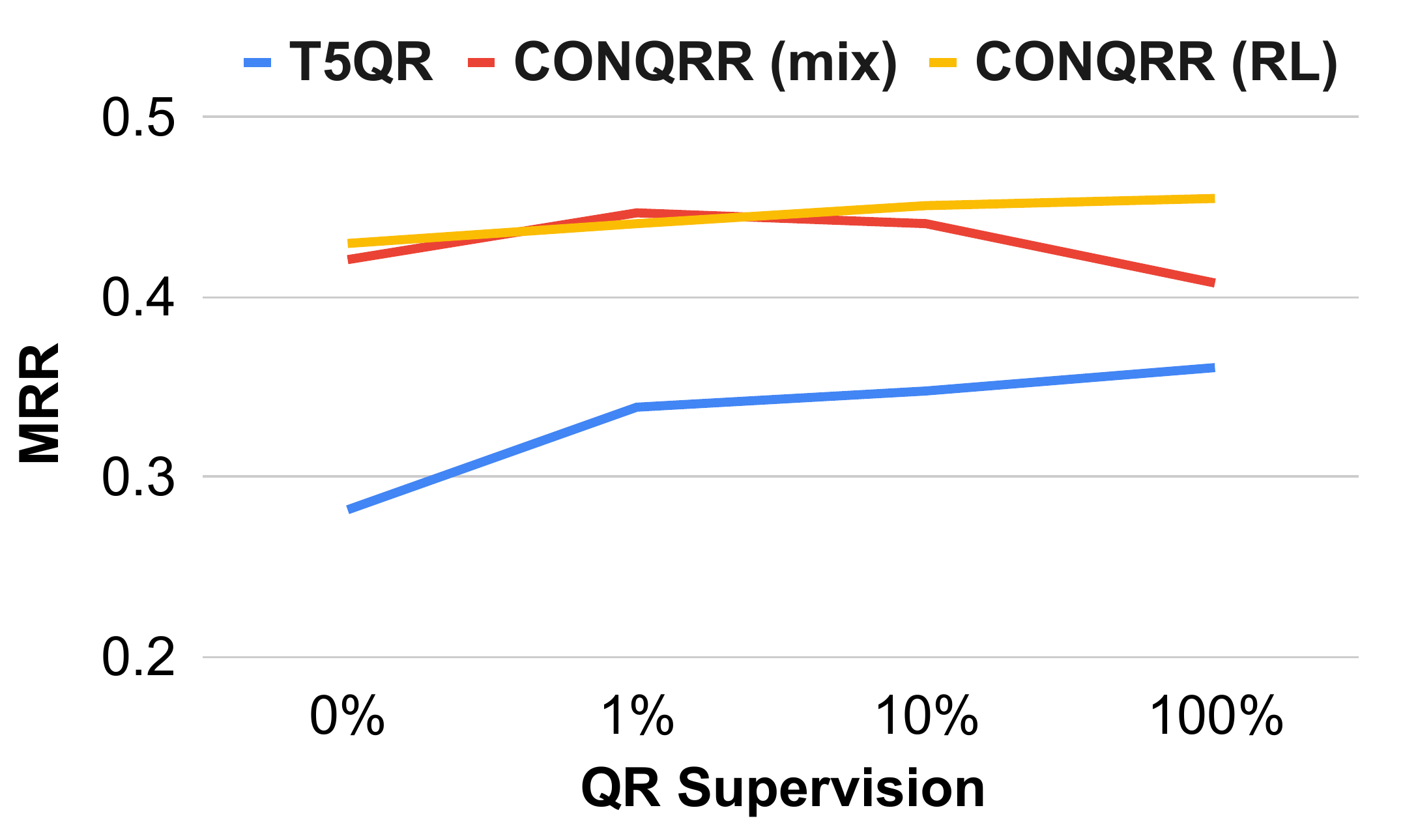}
  \caption{MRR on QReCC versus the percentage of QR supervision used for training, with DE as the retriever.}
 \label{fig:data_efficiency}
\end{figure}

\paragraph{Zero or Few QR Supervision} We investigate how sensitive \ours and T5QR are to the availability of QR labels. We experiment with training T5QR with 0\%, 1\%, 10\% or 100\% of QR labels in the QReCC train set. For the case of 0\% examples, we simply use the original T5 checkpoint without fine-tuning. When training \ours, we mask out the CE loss in Eq. (\ref{eq:loss}) for unused QR labels in training its initialized T5QR model, and we use the concatenated dialogue context as the BM25 input to obtain weak gold and hard negative passages for each training example, instead of using human rewrites (see details in Section~\ref{sec:model}). Figure~\ref{fig:data_efficiency} plots the curve of MRR on the overall QReCC test data using DE as the retriever versus the percentage of QR labels used for training. We see that \ours can already significantly outperform T5QR with even 0\% or 1\% of QR supervision. 

The slight difference in performance for the 100\% QR label case with respect to Table~\ref{tab:main_results} is due to the different mechanism (using human rewrite vs. the dialogue context) for choosing the positive and hard negative passages for RL training. Performance of the RL and mixed loss are similar when there is little supervision, roughly tracking the trends of the T5QR model that it is initialized with. The finding that performance degrades for the mixed loss with 100\% supervision may be due to a mismatch in the CE and RL losses as minimizing the CE loss does not directly optimize the retrieval performance.
T5QR is more sensitive to QR supervision but also does not require many QR labels for training, as its curve becomes flattened after 1\% supervision. We see similar trends with other metrics and BM25 (see Appendix~\ref{sec:appendix_analysis}).

\paragraph{Effects of Topic Shift \& Human Rewrites}
We hypothesize that a context involving a topic shift will present the greatest challenges for conversational passage retrieval. To explore this factor, we split the QReCC data into topic-concentrated and topic-shifted subsets as follows. A test example (with at least one previous turn) is considered
\textit{topic-concentrated} if the gold passage of the current question comes from a document that was used in \textit{at least one} previous turn. In contrast, a test example (with at least one previous turn) is considered \textit{topic-shifted} if the gold passage of the current question comes from a document that was \textit{never} used in any previous turn.
There are about 4.7k and 1.1k examples in the topic-concentrated and topic-shifted subsets, respectively. We compare the retrieval performance of different retriever inputs: dialogue context (which uses the concatenated dialogue history without QR), the predicted rewrite from T5QR and \ours with two loss alternatives, and the human rewrite.
Table~\ref{tab:dialogue_context} shows that the dialogue context outperforms even the human rewrite on the topic-concentrated set by 22\% and 17\%, averaging over three metrics, for BM25 and DE respectively, which shows the \textit{limitation of human rewrites}. We also see that \ours (RL) surpass the human rewrite on the topic-concentrated set on MRR for BM25 and all three metrics for DE.

\begin{table}
\centering
\resizebox{0.49\textwidth}{!}{
\begin{tabular}{ll|ccc|cccc}\toprule
\multirow{3}{*}{\textbf{Input}}
&\multirow{3}{*}{\textbf{IR}}
&\multicolumn{3}{c}{\textbf{Topic-Concentrated}}
&\multicolumn{3}{c}{\textbf{Topic-Shifted}}\\
 & & MRR & R10 & R100 & MRR & R10 & R100 \\\midrule
Dial Context & BM25  & \textbf{0.620} & \textbf{81.4} & \textbf{94.9} & 0.154 & 39.1 & 68.6 \\
T5QR & BM25 & 0.352 & 54.4 & 84.0 & \textbf{0.252} & 45.1 & 79.1 \\
\ours (mix) & BM25 & 0.419 & 63.1 & 91.2 & \textbf{0.252} & \textbf{45.9} & \textbf{82.1} \\
\ours (RL) & BM25 & 0.444 & 66.2 & 90.3 & 0.233 & 44.5 & 78.4 \\ \midrule
Human Rewrite & BM25 & 0.440 & 66.7 & 98.8 & 0.318 & 56.7 & 98.4 \\\midrule \midrule
Dial Context & DE  & \textbf{0.551} & \textbf{78.1} & \textbf{93.2} & 0.179 & 35.7 & 61.4 \\
T5QR & DE & 0.353 & 55.7 & 75.4 & 0.329 & 50.8 & 69.2 \\
\ours (mix) & DE & 0.404 & 63.8 & 83.4 & \textbf{0.334} & \textbf{53.2} & 72.6  \\
\ours (RL) & DE & 0.445 & 69.3 & 87.8 & 0.303 & 50.4 & \textbf{73.3}  \\ \midrule
Human Rewrite & DE & 0.424 & 65.5 & 84.5 & 0.397 & 61.0 & 79.8 \\ 
\bottomrule
\end{tabular}
}
\caption{Performance of using different retriever inputs for \textit{Topic-Concentrated} or \textit{Topic-Shifted} examples. }\label{tab:dialogue_context}
\end{table}

However, for the topic-shifted set, the human rewrite outperforms the dialogue context by 52\% and 61\%, averaging over three metrics, on BM25 and DE, respectively. The predicted rewrite by \ours (mix) outperforms the dialogue context by 30\% and 44\% on BM25 and DE, respectively. Therefore, compared with dialogue context, QR has great value in the aspect of \textit{robustness to topic shifts}. When comparing with human rewrites, we also see room for improvement for QR models.

These observations are \textit{largely unexplored} in previous work, and they motivate our work on the task of QR for conversational passage retrieval in general, and optimizing directly towards retrieval. 
\begin{figure}
  \centering
  \includegraphics[width=0.49\textwidth]{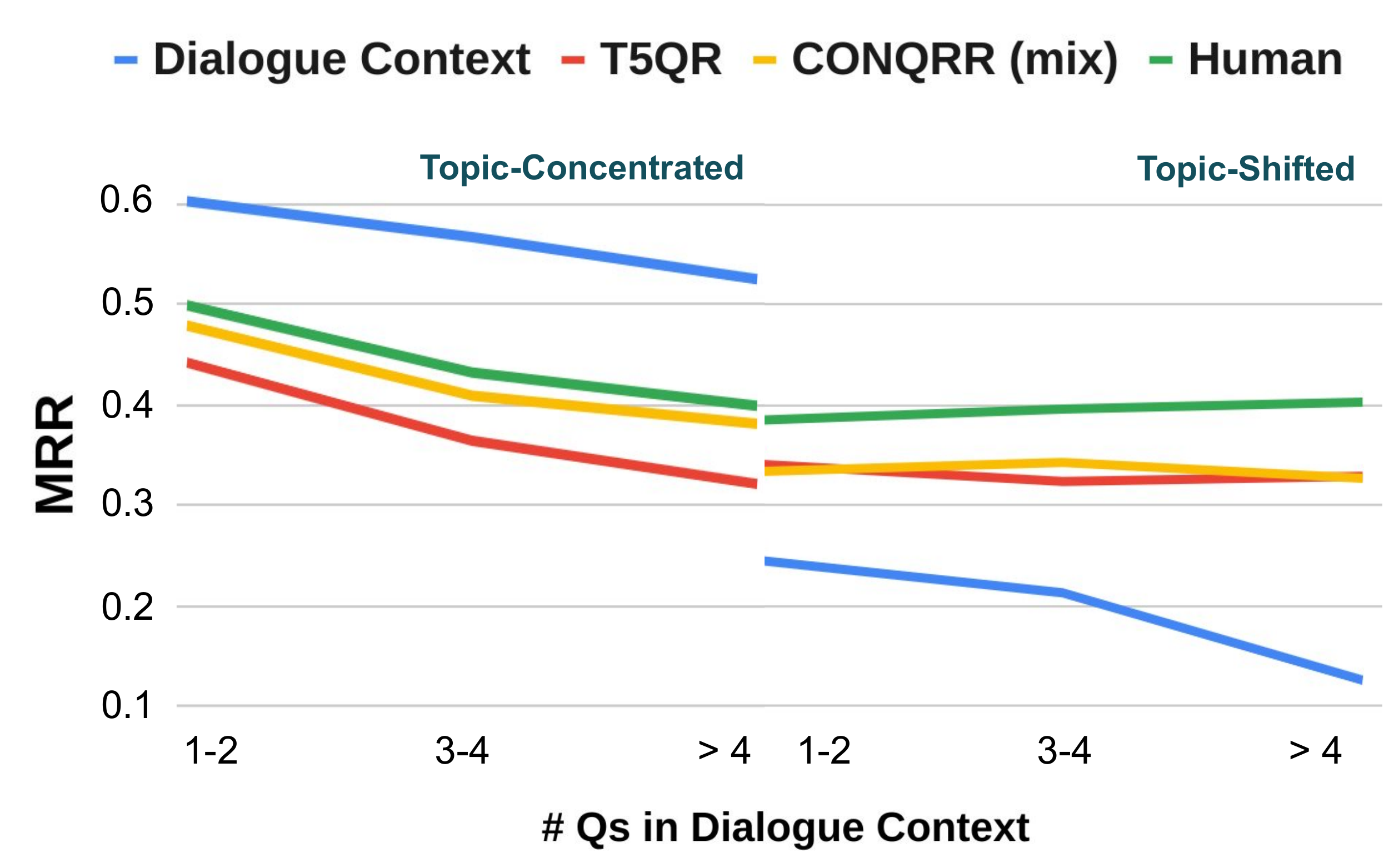}
  \caption{MRR versus the number of questions in the dialogue context, with DE as the retriever.}
 \label{fig:recall_vs_ctx_length}
\end{figure}

\begin{table*}
\centering
\small
\resizebox{1.0\textwidth}{!}{
\begin{tabular}{p{1.2cm}|p{7.5cm}|p{8.4cm}}
\toprule
Dialogue Context & \textit{Q}: What were \textbf{John Stossel}'s most popular publications? \newline
\textit{A}: Give Me a Break: How I Exposed Hucksters, Cheats, and Scam Artists and Became \dots \newline
\dots \newline
\textit{Q}: What was the response? & 
\textit{Q}: What were some notable live performances at the Buena Vista Social Club? \newline
\textit{A}: \textbf{Ibrahim Ferrer and Rubén González} \dots \newline
\dots \newline
\textit{Q}: What other live performances are important? \\
\midrule
Gold \newline Passage & \textbf{Stossel} has written three books. Give Me a Break: \dots It was \textit{a New York Times bestseller for 11 weeks} \dots & The first performances \dots \textbf{Ibrahim Ferrer and Rubén González} performed together \dots \textit{a 1999 Miami performance} \dots \\ \midrule
\ours (mix) & What was the response to \textbf{John Stossel}'s book, Give Me a Break? (Rank=2) & What other live performances at the Buena Vista Social Club are important besides \textbf{Ibrahim Ferrer and Rubén González}? (Rank=2) \\
T5QR & What was the response to the book Give Me a Break? (Rank >100) & What other live performances are important at the Buena Vista Social Club? (Rank=18) \\
Human & What was the response to Give Me a Break: How I Exposed Hucksters, Cheats, and Scam Artists and Became the Scourge of the Liberal Media?
 (Rank >100) & What other live performances of the Buena Vista Social Club are important? (Rank=17) \\
\bottomrule
\end{tabular}
}
\caption{\label{tab:examples} Examples of predicted rewrites and the gold passage ranks by using them as the DE retriever input. \textit{The gold answer is italicized in the gold passage.}
}
\vspace{-1mm}
\end{table*}
\begin{table}
\centering
\small
\resizebox{0.49\textwidth}{!}{
\begin{tabular}{lcc|cc|ccc}\toprule
\multirow{3}{*}{\textbf{QR Model}}
&\multicolumn{2}{c}{\textbf{QuAC-Conv}}
&\multicolumn{2}{c}{\textbf{NQ-Conv}} &\multicolumn{2}{c}{\textbf{TREC-Conv}}\\
&L & \% OL &L & \% OL &L & \% OL \\\midrule
T5QR                & 10.9 & 35.8 & 8.9 & 40.4 & 8.2 & 37.8 \\
Ours (mix) w/ BM25  & 12.1 & 37.2 & 9.5 & 42.1 & 8.5 & 38.8  \\
Ours (RL) w/ BM25   & 11.2 & 40.2 & 10.1 & 44.6 & 9.4 & 39.4 \\
Ours (mix) w/ DE    & 12.1 & 37.2 & 9.6 & 41.7 & 8.7 & 39.1 \\
Ours (RL) w/ DE     & 28.2 & 51.1 & 21.7 & 55.8 & 18.3 & 44.3 \\\midrule
Human               & 12.1 & 37.2 & 9.3 & 43.0 & 8.4 & 41.7 \\
\bottomrule
\end{tabular}
}
\caption{Average number of tokens (L) and the percentage of overlapping tokens (OL) with the gold passage(s) in output rewrites. 
}
\label{tab:rewrite_stats}
\vspace{-2mm}
\end{table}

\vspace{-1mm}
\paragraph{Effect of Dialogue Context Length} 
Figure~\ref{fig:recall_vs_ctx_length} shows the MRR score on topic-concentrated and topic-shifted subsets with DE as the retriever for various dialogue context lengths.
Dialogue context lengths are grouped into 1-2, 3-4 and $\ge4$ previous utterances (including the current question). For topic-concentrated conversations, all compared models have similar robustness to the dialogue context length and \ours (mix) is slightly more robust than T5QR. For topic-shifted conversations, both QR models and human rewrites show little drop or even an increase in performance as the context length gets longer. In contrast, the robustness of the dialogue context worsens with longer contexts, which confirms the importance of QR discussed above. We have similar observations for other metrics as well as for the BM25 retriever.

\vspace{-1mm}
\paragraph{Quantitative Attributes of Rewrites} Table~\ref{tab:rewrite_stats} shows the average number of tokens per rewrite, and the percentage of overlapping tokens (excluding stopwords) between the rewrite and the gold passage(s). \ours generally generates longer rewrites with more overlapping tokens with gold passage(s), compared with T5QR. With DE as the retriever, \ours (RL) generates more than double the length of T5QR, \ours (mix) and even human rewrites. We show in Appendix~\ref{sec:appendix_analysis} that T5QR underperforms \ours even when we make it generate rewrites of similar lengths by applying a brevity penalty \citep{wu2016googles}.

\paragraph{Rewrite Quality Analysis and Examples} 
In order to understand why rewrites generated by \ours lead to better retrieval performance and even sometimes outperform human rewrites,\footnote{This is only for analysis purposes. Note that the goal of our predicted rewrites is to improve retrieval performance instead of directly being used by end users. } we sampled 50 examples where \ours (mix) leads to better ranking of gold passages than human rewrites (using DE retriever).  We notice that 70\% of \ours generated rewrites contain additional context and (correct) information when compared to human rewrites. The remaining 30\% contain alternative or less context information than human rewrites. In such cases, potentially because the information in human rewrites is less relevant to gold passages, it led to a lower gold passage rank. Overall, these \ours rewrites are as fluent as human rewrites and contain no major misinterpretation of the dialogue context.
Table~\ref{tab:examples} shows two examples of generated rewritten queries of T5QR and \ours (mix) trained with DE in the loop, as well as the human rewrites. In the left example, the \ours rewrite includes an entity ``John Stossel'' that is mentioned in the gold passage but not included by rewrites from T5QR or Human. Thus, even if the human rewrite is longer by containing the book's full name, \ours enables more efficient retrieval with a partial book name along with its author name. In the right example, \ours generates a longer rewrite containing richer contextual information. We have similar observations for BM25 and put more examples in Appendix~\ref{sec:appendix_analysis}. 
 
For error analysis, we sampled another 50 examples where \ours (mix) leads to worse ranking of gold passages than human rewrites with DE. All were deemed fluent. We found in most of these cases, \ours rewrites contain less context than human rewrites (56\%) or additional information with a misinterpretation of the user request (34\%). See examples in Appendix~\ref{sec:appendix_analysis} due to space limit. 


\section{Conclusion and Future Work}
\label{sec:conclusion}
To summarize, we introduce \ours to address query rewriting for conversational passage retrieval with an off-the-shelf retriever. Motivated by our analysis showing both the limitations and utility of human rewrites, which are unexplored by prior work, we adopt RL with a novel reward to train \ours directly towards retrieval. As shown, \ours is the first QR model that can be trained adaptively to any off-the-shelf retriever, and achieves state-of-the-art retrieval performance on QReCC with conversations from 3 different sources. It shows better performance with zero QR supervision when compared with strong supervised baselines trained with full QR supervision.

A direction for future work includes leveraging QR to facilitate other tasks like question answering and response generation in a full CQA system, as well as sentence rewriting in a document \citep{10.1162/tacl_a_00377}. Future investigation is needed to explore conversations with other discourse relations like asking for clarifications besides alternating questions and answers in current CQA datasets.

\section*{Limitations}
We show in Section~\ref{sec:analysis} (Table~\ref{tab:dialogue_context}) that compared to directly use dialogue context without QR, a QR model has great value in robustness to topic shifts when used with an off-the-shelf retriever. 
However, if most conversations of interest are topic-concentrated, we show that using the dialogue context itself may already work well. Although we focus on the \textit{fixed retriever} setting in this work,  we illustrate in Table~\ref{tab:finetune_de} in Appendix~\ref{sec:appendix_analysis}, that if the downstream retriever is \textit{allowed to be fine-tuned}, our best QR model \ours (mix) underperforms compared to the dialogue context in both topic-concentrated and topic-shifted scenarios, and thus the benefits of QR as an intermediate step require further justification in that setting. Nevertheless, the table still shows that human rewrites have an advantage on topic-shifted conversations over dialogue contexts. Therefore, it would be interesting for follow-up studies to investigate the design of a QR model that reaches close performance with human rewrites on topic-shift scenarios with a fine-tunable retriever. Then, combining the dialogue context with the rewritten query for retrieval may help further improve the overall retrieval performance. 

The training time of \ours is longer than fine-tuning a DE retriever of a similar model size (9 vs 2 hours) because for each training step of \ours, \ours needs to do autoregressive decoding to get greedily decoded and sampled $q$ and $q_s$. However, re-indexing passages after fine-tuning the retriever can be very time-consuming (about 24 hours) and memory-consuming. In addition, unlike DE, \ours can also be used for any blackbox retriever such as search engines that are infeasible to fine-tune or be replaced.

Another downside of QR is that for out-of-domain and topic-shifted scenarios, QR may still require additional labels to achieve robust performance. Although we show that \ours (RL) initialized with T5 does not require QR labels and can work well on the overall QReCC test set, \ours (RL) does show worse robustness to out-of-domain and topic-shifted examples when compared with \ours (mix). Therefore, training a more robust \ours model may still require additional annotation efforts to collect human rewrites. 

\ours has only been tested on the standard CQA dialogue format of alternating questions and answers. To facilitate more practical use cases with more diverse dialogue acts or discourse relations (e.g., the agent asks a clarification question to the user), further investigation is needed.

\section*{Ethical Considerations}
\label{sec:ethical}
Our work is primarily intended to leverage query rewriting (QR) models to facilitate the task of conversational passage retrieval in an open-domain CQA system. Retrieving the most relevant passage(s) to the current user query in a conversation would help to generate a more appropriate agent response. Predicted rewrites from our QR model are mainly intended to be used as \textit{intermediate} results (e.g., the inputs to the downstream retrieval system). 
They may also be useful for interpretability purposes when a final response does not make sense to the user in a full CQA system, but that introduces a potential risk of offensive text generation.
In addition, to prevent the retriever from retrieving passages from unreliable resources, filtering of such passages in the corpus should be performed before any practical use.

\section*{Acknowledgements}
We thank all the Google AI Language and UW NLP members who provided help to this work. We specifically want to thank Kristina Toutanova, Ming-Wei Chang, Kenton Lee, Daniel Andor, Matthew Lamm, Victoria Fossum, Iulia Turc, Dipanjan Das and Elizabeth Clark for their valuable insights and feedback. We thank Zhuyun Dai and Vincent Zhao for their input in early discussions, and Jianmo Ni for his help on setting up the dual encoder model. We also would like to thank all anonymous reviewers for providing detailed and insightful comments on our work.

\bibliography{emnlp2022}
\bibliographystyle{acl_natbib}

\appendix
\newpage
\section{Appendix}
\label{sec:appendix}

\subsection{Additional Implementation Details}
\label{sec:appendix_impl}
Our models are implemented using JAX.\footnote{\url{https://github.com/google/jax}}
For training, we set 64, 1k and 10k as the batch size, warm-up steps and total training steps, respectively. We use $e^{-3}$ and $e^{-4}$ as the learning rate for T5QR and \ours, respectively. We use Adafactor \citep{pmlr-v80-shazeer18a} as our optimizer with the default parameters. Linear decay is applied after 10\%
of the total number of training steps, reducing the
learning rate to 0 by the end of training. For supervised training, models are selected based on the best dev set Rouge-1 F1 score with the human rewrites, following \citet{anantha-etal-2021-open}. 
For RL-based training of \ours, models are selected based on the average in-batch gold passage prediction accuracy as in Eq. (\ref{eq:score}) on dev set with greedily decoded rewrites. 
For the experiment with the pure RL loss and the retriever BM25, our results are obtained with the initialized T5QR model being fine-tuned with only 10\% QR labels, as we find initializing with a model using 100\% QR labels is unstable for BM25.
Previous work \citep{wu-etal-2021-automatic} also had a similar observation that initializing with a less trained model leads to more stable RL training. 

The maximum length of the dialogue context fed into the QR model is 384 (longer than 97.9\% dialogue contexts in QReCC) and the maximum output rewrite length is 64 (longer than 99.9\% human rewrites). To generate each sampled rewrite $q_s$ (see Section~\ref{sec:model}), we apply top-k sampling where $k=20$. For each training example, we sample 5 rewrites in total (i.e., $m=5$ for the RL training explained in Section~\ref{sec:model}). Each training process is run on 8 TPU nodes. It takes about 2 and 9 hours for the supervised and RL-based training, respectively. For each experiment, we observe similar performance or training curves for 2-3 runs and report numbers on a random run. Both T5QR and \ours are based on T5-base and have about 220M parameters. In contrast, the baseline Transformer++ is based on GPT2-medium and has about 345M parameters.

For the BM25 retriever model, Pyserini \cite{yang-anserini} is used with defaults $k_1=0.82$, $b=0.68$. These values were chosen based on retrieval performance on MS MARCO \cite{nguyen2016ms}, which contains non-conversational queries only. During the RL training of \ours, due to the complexity of applying Pyserini to calculate rewards on-the-fly, we instead use a Pyserini approximate called BM25-light. The only differences between them are that BM25-light (1) uses T5’s subword tokenization instead of whole word tokenization and (2) does not use special operations (e.g., stemming) as applied in Pyserini. After training, we still run inference and report retrieval performance on BM25. Pyserini simply encodes the whole query input and each passage without truncating. We set maximum query and passage length as 128 and 2000 for BM25-light, but only less than 0.1\% cases require truncation with these thresholds. 

For the dual encoder, the maximum query or passage length is 384. The average passage length is 378, but we observe performance drop by further increasing the maximum length for the dual encoder.

\subsection{Additional Data Details}
\label{sec:appendix_data}
QReCC reuses questions in QuAC and TREC conversations and re-annotates answers. For each NQ-based conversation, they only use one randomly chosen question from NQ to be the starting question and then annotate the remaining conversation.
In total, there are 63k, 16k and 748 question and answer pairs in the three subsets QuAC-Conv, NQ-Conv, TREC-Conv respectively, where TREC-Conv only appears in the test set. The original data is only divided into train and test sets. We randomly choose 5\% examples from the train set to be our validation set.

In some conversations from QuAC-Conv, the first user query is ambiguous as it depends on some topical information from the original QuAC dataset. Therefore, in order to fix this issue, we follow \citet{anantha-etal-2021-open} to replace all first user queries in QReCC conversations with the their corresponding human rewrites.

QReCC is a publicly available dataset that was released under the Apache License 2.0 and we use the same task set-up proposed by the original QReCC authors.

\subsection{Additional Evaluation Details}
\label{sec:appendix_eval}
Some agent turns in QReCC do not have valid gold passage labels,\footnote{Missing gold labels for certain examples in the dataset has no effect on the training of \ours as we induce weak labels without using the provided labels.} and the (provided) original evaluation script assigns a score of 0 to all such examples. Their updated evaluation script calculates the scores by removing those examples from the evaluation set (roughly 50\%), which results in 6396, 1442 and 371 test instances for QuAC-Conv, NQ-Conv and TREC-Conv, respectively. 
This leads to a total of 8209 test instances in QReCC. 
We use the \textit{updated} evaluation script for most of our experiments, except that we also use the \textit{original} version for calculating scores in Table~\ref{tab:comparable_eval} to compare with their reported QReCC baseline results. We note that these two evaluation scripts only differ by a scaling factor so they should lead to the same conclusions regarding model comparisons.

\subsection{Additional Analysis}
\label{sec:appendix_analysis}

\paragraph{Lower Recall@100 with DE} Previous work \citep{karpukhin-etal-2020-dense} shows that DE retrievers generally lead to better recall scores than BM25. However, in Table~\ref{tab:main_results}, we observe that across all subsets, the best MRR and Recall@10 results are consistently from DE, whereas BM25 has better Recall@100 scores. One reason to explain the observation difference is that we use an \textit{off-the-shelf} retriever for our retrieval task while most previous work that compares BM25 and DE focuses on fine-tuning the DE model. Without being fine-tuned, a DE model may be more vulnerable to domain shift than BM25. On the other hand, prior work \citep{luan-etal-2021-sparse} proves that a DE model's performance would drop as the passage length increases. In the QReCC dataset, the average passage length is 378, which is relatively long \citep{luan-etal-2021-sparse}. 

\begin{figure}[!htbp]
  \centering
  \includegraphics[width=0.45\textwidth]{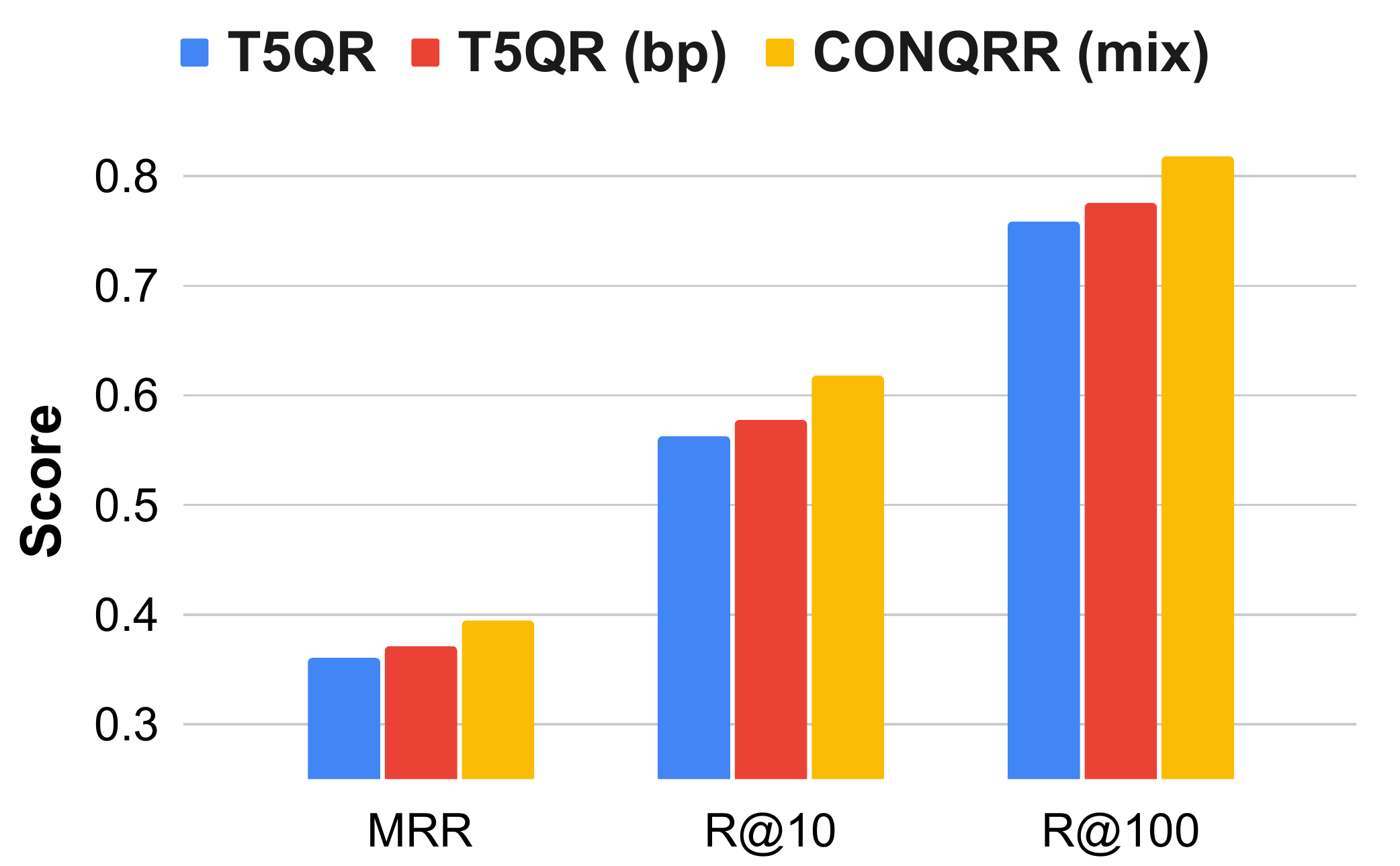}
  \caption{Evaluation scores on QReCC for T5QR w/ or w/o brevity penalty and \ours (mix), with DE as the retriever. Recall scores (R@k) are divided by 100.}
 \label{fig:bp}
\end{figure}

\paragraph{Analysis of Longer Rewrites} We hypothesize that simply generating a longer rewritten query is not the only factor that contributes to better retrieval performance. We investigate this by applying a brevity penalty \citep{wu2016googles} during decoding for T5QR such that its average query length matches that of \ours (mix). Figure~\ref{fig:bp} shows that \ours (mix) still outperforms T5QR with the brevity penalty for all three evaluation metrics on QReCC.

\paragraph{Fine-tuned Retriever} Although our work focuses on the off-the-shelf retriever setting, we also conduct an experiment of fine-tuning the DE retriever with the concatenated dialogue context, the predicted rewrite from \ours (mix) or the human rewrite as the query input, with results in Table~\ref{tab:finetune_de}. The numbers are comparable to those in Table~\ref{tab:dialogue_context}. 
Fine-tuning the DE retriever improves results for all scenarios, but the dialogue context benefits substantially, to the extent that it outperforms ConQRR in topic-shifted cases. However, there is still improvement room as we see benefits of human query-rewrites for topic shifts. 
\begin{table}
\centering
\resizebox{0.49\textwidth}{!}{
\begin{tabular}{l|ccc|cccc}\toprule
\multirow{3}{*}{\textbf{Input}}
&\multicolumn{3}{c}{\textbf{Topic-Concentrated}}
&\multicolumn{3}{c}{\textbf{Topic-Shifted}}\\
 & MRR & R10 & R100 & MRR & R10 & R100 \\\midrule
Dial Context & \textbf{0.643} & \textbf{87.7} & \textbf{96.9} & \textbf{0.312} & \textbf{56.2} & \textbf{81.9} \\
\ours (mix) & 0.588 & 84.0 & 96.9 & 0.259 & 48.3 & 77.2 \\\midrule
Human Rewrite & 0.510 & 79.9 & 95.2 & 0.380 & 61.3 & 86.0 \\
\bottomrule
\end{tabular}
}
\caption{Results of using the dialogue context, predicted rewrite or human rewrite as the retriever input with the \textit{finetuned} DE as the retriever. }\label{tab:finetune_de}
\end{table}

\begin{figure}
  \centering
  \includegraphics[width=0.45\textwidth]{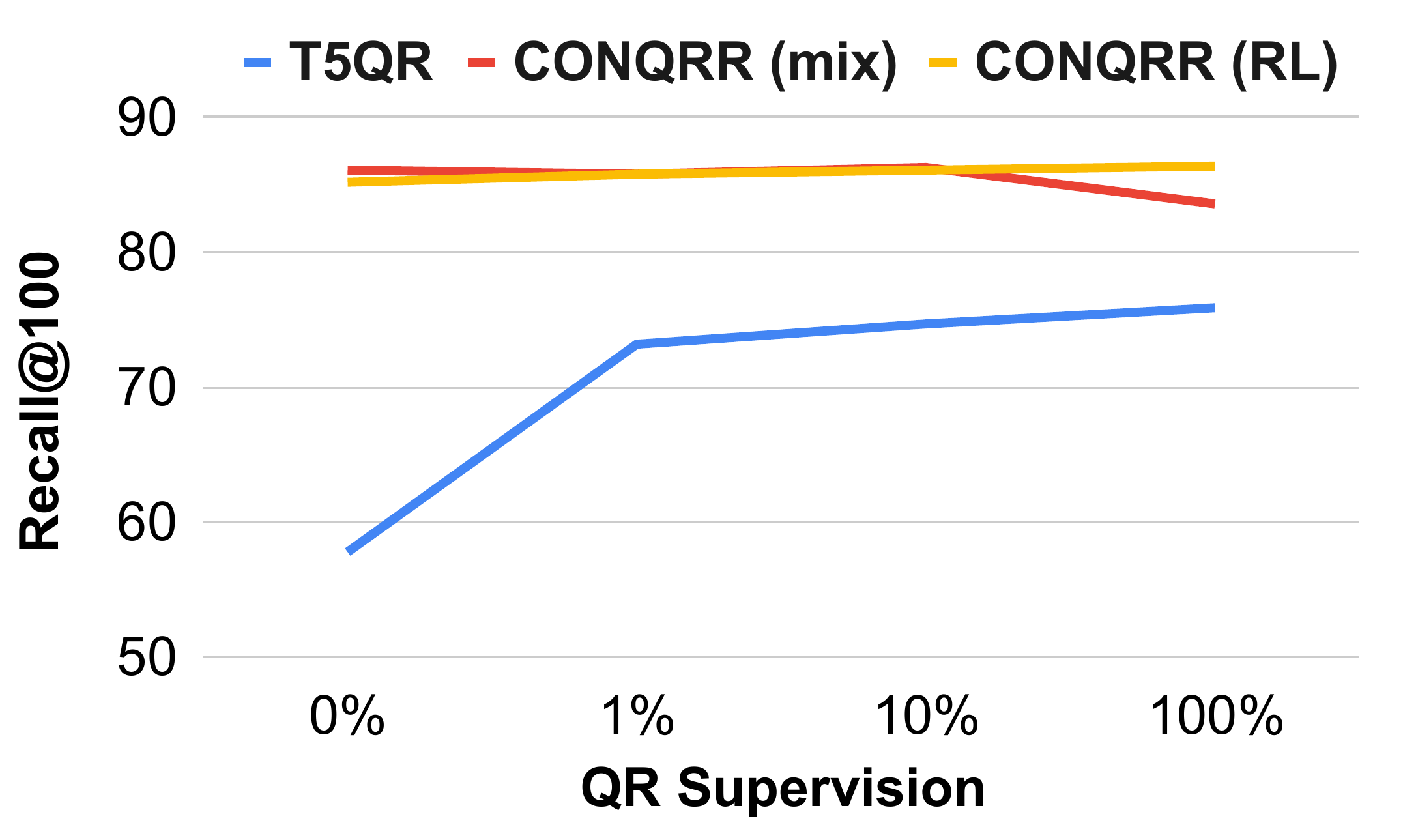}
  \caption{Recall@100 on QReCC versus the percentage of QR supervision used for training, with DE as the retriever.}
 \label{fig:data_efficiency_r100}
\end{figure}

\paragraph{Additional Data Efficiency Figure} Figure~\ref{fig:data_efficiency_r100} shows the curve of Recall@100 on the overall QReCC test data using DE as the retriever versus the percentage of QR labels used for training. We also observe similar trends with Recall@10 and using BM25 as the retriever.

\begin{table*}[h]
\centering
\small
\resizebox{1.0\textwidth}{!}{
\begin{tabular}{p{1.2cm}|p{5.7cm}|p{9.7cm}}
\toprule
Dialogue Context & \textit{Q}: How did Michael Anthony's career start? \newline
\textit{A}: While attending Pasadena City College, Anthony met Eddie Van Halen \dots Bassist Mark Stone left Mammoth. \newline
\textit{Q}: How was that band formed? & 
\textit{Q}: What kind of instrumentation did Pink Floyd use on the album The Dark Side of the Moon? \newline
\dots \newline
\textit{Q}: Were there any particular songs they used this technique on?\newline
\textit{A}: Speak to Me and Money. \newline
\textit{Q}: What other different techniques did they use? \\
\midrule
Gold \newline Passage &  \textbf{Anthony} met \dots \textit{Van Halens decided to audition Anthony as a replacement. Anthony was impressed by their skill during subsequent jam sessions even though he had seen the brothers play before} \dots & The album features metronomic sound effects \dots \textit{The sound effects on ``Money'' were created by splicing together Waters' recordings of clinking coins, tearing paper, a ringing cash register, and a clicking adding machine} \dots Pink Floyd \dots \\ \midrule
\ours (mix) & How was the band Mammoth formed by \textbf{Michael Anthony}? (Rank=0) & What other different techniques did Pink Floyd use besides metronomic sound effects and tape loops? (Rank=4) \\
T5QR & How was the band formed? (Rank >100) & What other different techniques did Pink Floyd use on the album The Dark Side of the Moon besides metronomic sound effects and tape loops? (Rank=55) \\
Human & How was Mammoth formed after Mark Stone left Mammoth? (Rank=31) & What other different techniques did Pink Floyd use on the album The Dark Side of the Moon besides metronomic sound effects and tape loops? (Rank=55) \\
\bottomrule
\end{tabular}
}
\caption{\label{tab:appendix_examples_de} Additional Examples of predicted rewrites and the gold passage ranks by using them as the \textbf{DE retriever} input. In these examples, \ours predicts alternative or less context information than human rewrites, but leads to a lower gold passage rank. \textit{The gold answer is italicized in the gold passage.}}

\end{table*}

\begin{table*}[h]
\centering
\small
\resizebox{1.0\textwidth}{!}{
\begin{tabular}{p{1.2cm}|p{7.7cm}|p{7.7cm}}
\toprule
Dialogue Context & \textit{Q}: What is Get 'Em Girls? \newline
\textit{A}: Jessica Mauboy's second studio album, \textbf{Get 'Em Girls} (\textbf{2010}). \newline
\dots \newline
\textit{Q}: Did she receive any awards or honors during these years? & 
\textit{Q}: What is one actress who was a Bond girl? \newline
\textit{A}: Ursula Andress in \textbf{Dr. No} is widely regarded as the first Bond girl. \dots \newline
\dots \newline
\textit{Q}: Who was another Bond girl? \\
\midrule
Gold \newline Passage & \dots Mauboy performed ``\textbf{Get 'Em Girls}'' at the \textbf{2010} \dots received \textit{her first nomination for Young Australian of the Year} \dots & \dots Ursula Andress (as Honey Ryder) in \textbf{Dr. No} (1962) is widely regarded as the first Bond girl, although she was preceded by both \textit{Eunice Gayson} as Sylvia Trench and \dots \\ \midrule
\ours (mix) & Did Jessica Mauboy receive any awards or honors during the years she released \textbf{Get 'Em Girls}? (Rank=7) & Who was another Bond girl besides Ursula Andress in \textbf{Dr. No}? (Rank=7) \\
T5QR & Did Jessica Mauboy receive any awards or honors during these years? (Rank >100) & Who was another Bond girl? (Rank=68) \\
Human & Did Jessica Mauboy receive any awards or honors during the \textbf{2010}s? (Rank=24) & Who was another Bond girl, besides Ursula Andress? (Rank=12) \\
\bottomrule
\end{tabular}
}
\caption{\label{tab:appendix_examples_bm25} Examples of predicted rewrites and the gold passage ranks by using them as the \textbf{BM25 retriever} input. \textit{The gold answer is italicized in the gold passage.}}
\end{table*}

\begin{table}
\centering
\small
\resizebox{0.49\textwidth}{!}{
\begin{tabular}{p{1.2cm}|p{7.7cm}}
\toprule
Dialogue Context & \textit{Q}: What did Jan Howard do in the early 60s? \newline
\textit{A}: In 1960, Jan Howard went to Nashville, Tennessee, where they appeared on The Prince Albert Show, the Grand Ole Opry segment carried nationally by NBC Radio. \newline
\textit{Q}: Did she get a record deal? \\
\midrule
\ours (mix) & Did Jan Howard get a record deal? (Rank=69)\\
Human & Did Jan Howard get a record deal in 1960 after her appearance on The Prince Albert Show? (Rank=6)\\
\bottomrule
\end{tabular}
}
\caption{\label{tab:appendix_error_example_less_info} \textbf{Error analysis example 1}: \ours (mix) rewrite contains less context than the human rewrite, which leads to worse ranking of the gold passage.}
\end{table}

\begin{table}
\centering
\small
\resizebox{0.49\textwidth}{!}{
\begin{tabular}{p{1.2cm}|p{7.7cm}}
\toprule
Dialogue Context & \textit{Q}: What is the keto diet?\newline
\dots \newline
\textit{A}: The Paleolithic diet, Paleo diet, caveman diet, or stone-age diet is a modern fad diet requiring the sole or predominant eating of foods presumed to have been available to humans during the Paleolithic era. \newline
\textit{Q}: What do they have in common?\\
\midrule
\ours (mix) & What do the Paleolithic diet and the stone-age diet have in common? (Rank=78)\\
Human & What do paleo diet and keto diet have in common? (Rank=1)\\
\bottomrule
\end{tabular}
}
\caption{\label{tab:appendix_error_example_misinterpred} \textbf{Error analysis example 2}: \ours (mix) rewrite contains a misinterpretation of the user request, which leads to worse ranking of the gold passage than the human rewrite.}
\end{table}

\paragraph{Additional Rewrite Examples} In addition to Table~\ref{tab:examples}, we put more examples in Table~\ref{tab:appendix_examples_de} for using DE as the retriever. We also put predicted rewrites from \ours (mix) that is trained towards BM25 instead of the DE retriever in Table~\ref{tab:appendix_examples_bm25}. Gold passage ranks are shown in the table, using the predicted rewrites as the BM25 retriever input. 

Table~\ref{tab:appendix_error_example_less_info} and~\ref{tab:appendix_error_example_misinterpred} contain examples where CONQRR (mix) rewrites have worse ranking of the gold passage than human rewrites, from our error analysis. In the two examples, the \ours rewrite contains less context than human rewrites or a misinterpretation of the user request.


\end{document}